\pdfoutput=1

\documentclass[11pt]{article}

\usepackage{emnlp2022}

\usepackage{times}
\usepackage{latexsym}
\usepackage{amsmath}
\usepackage{amssymb}
\usepackage{tabularx}
\usepackage{xcolor}
\usepackage{booktabs}
\usepackage{graphicx}
\usepackage{multirow}

\usepackage[T1]{fontenc}

\usepackage[utf8]{inputenc}

\usepackage{microtype}

\title{Towards Corpus-Scale Discovery of Selection Biases in News Coverage: \\ Comparing What Sources Say About Entities as a Start}

\author{Sihao Chen \hspace{1cm} William Bruno \hspace{1cm} Dan Roth \\
University of Pennsylvania \\
\small \texttt{\{sihaoc, wwbruno, danroth\}@cis.upenn.edu}}

\begin{document}
\maketitle
\begin{abstract}
News sources undergo the process of selecting newsworthy information when covering a certain topic. 
The process inevitably exhibits \emph{selection biases}, i.e. news sources' typical patterns of choosing what information to include in news coverage, due to their agenda differences. To understand the magnitude and implications of selection biases, one must first discover (1) on what topics do sources typically have diverging definitions of "newsworthy" information, and (2) do the content selection patterns correlate with certain attributes of the news sources, e.g. ideological leaning, etc.   

The goal of the paper is to investigate and discuss the challenges of building scalable NLP systems for discovering
patterns of media selection biases directly from news content in massive-scale news corpora, without relying on labeled data. 
To facilitate research in this domain, we propose and study a conceptual framework, where we compare how sources typically mention certain controversial entities, and use such as indicators for the sources' content selection preferences. We empirically show the capabilities of the framework through a case study on NELA-2020, a corpus of $1.8$M news articles in English from 519 news sources worldwide. We demonstrate an unsupervised representation learning method to capture the selection preferences for how sources typically mention controversial entities. Our experiments show that that distributional divergence of such representations, when studied collectively across entities and news sources, serve as good indicators for an individual source's ideological leaning.  We hope our findings will provide insights for future research on media selection biases.

\end{abstract}

\begin{figure}[t!]
\centering
\includegraphics[width=0.95\linewidth]{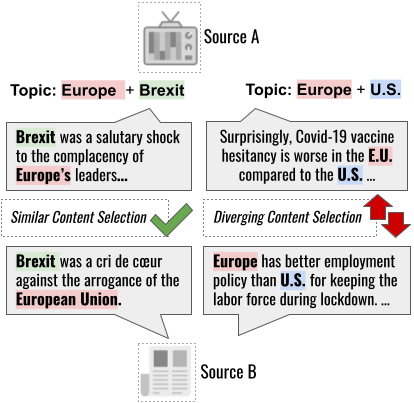}
\caption{Overview of our conceptual framework: We represent a sources' content selection preferences by comparing the content of news coverage on each fine-grained topic against other sources. We found that a source's selection preferences on different topics collectively serve as good indicators for its ideological leaning ($\S$~\ref{ssec:understand}). In our study, we choose to represent the topic of a text segment by the entities it mentions. 
}
\label{fig:lead_example}
\end{figure}

\section{Introduction}
News media represents one of the most prominent information channels to our eventful world. Although we are blessed with a wide variety of news sources, each source naturally has to go through the process of sampling and condensing information to include in news coverage. The editorial decisions of what is newsworthy or not directly influence the perceived representation of the reported topic or event by the audience \cite{d2000media, groeling2013media}. Such problem is generally referred to as \textit{media selection bias}\footnote{As \emph{selection bias} is a much overloaded term, we will define the precise scope of our study in $\S$~\ref{sec:background}.}. Understanding the patterns of media selection biases is crucial, yet intrinsically difficult. As the editorial decisions of various sources have different motivations, and their content selection preferences vary drastically by topic, it would require a holistic, massive scale of content analysis before we can uncover the magnitude and form of selection biases that each source exhibits.

In this paper, we discuss and investigate the challenges of developing scalable computational methods for discovering patterns of selection biases from news content directly. Our study revolves around two research questions. First, how do we represent a sources' selection preferences by each topic of the news content they produce? And subsequently, can we try to explain such selection preferences by quantifying their correlation with a set of the source's motivating attributes, such as ideological leaning?

We start off by discussing the challenges and limitations of existing methods from the NLP perspective ($\S$~\ref{sec:background}). Surrounding the two research questions, we propose and sketch a conceptual framework which uses \emph{named entities} to represent the \emph{topic} of a text segment.
Figure~\ref{fig:lead_example} illustrates our idea with a motivating example. 
From news articles published by different sources within similar time frames, we extract and categorize text segments by which two\footnote{We use two instead one, as we observe two entities in a sentence would provide focused and much finer-grained topic guidance, as shown in Figure~\ref{fig:lead_example}.} of the common named entities that they mention. When looking at such text segments collectively and comparatively across news sources, the aspect or context in which the two entities get mentioned would implicitly reflect the two sources' similar or different content preferences on the topic. We formulate the task of recognizing whether two text segments present similar context as an unsupervised representation learning objective, as we will describe in $\S$~\ref{ssec:oprl}. We observe that such entity-centric formulations provide a flexible way of aligning different sources' statements made on similar topics, and thus the similarity between those statements would conceptually reflect the similarity between the two sources' content selection preferences with respect to the topic. 

We assess the utility of the idea by conducting a case study on the NELA-2020 corpus \cite{horne2018sampling, gruppi2021nela}, which consists of $\tilde~1.8$M news articles in 2020 from 519 news sources worldwide. We first identify a set of most frequent entities through entity linking and corpus statistics as our topics of interest for the study ($\S$~\ref{ssec:preprocess}). With political bias ratings of the sources from media watchdog \emph{AllSides}\footnote{\url{https://www.allsides.com/media-bias/media-bias-ratings}}, we discover empirical evidence that the collective set of content selection preferences w.r.t. topics guided by entities show positive correlation with the ideological polarization among the sources.
Interestingly, we also observe that our methodology can be used to identify common grounds among sources, where sources typically agree with each other regardless of their ideological leaning.
Following such observation, we describe a statistical way to define and discover \emph{archetypical entity pairs}, for which sources with conflicting ideological preferences display polarized content selection preferences. We empirically show that archetypical entity pairs provide a straightforward way of recognizing the ideological leaning of a source.

In summary, our contributions are three-fold:
\begin{enumerate}
    \item We investigate the challenges of discovering patterns of media selection bias from massive-scale news corpora. We discuss and connect the research questions with the goal of building scalable NLP systems for such.   
    \item We propose a conceptual framework for the challenge, which uses \emph{named entities} to represent the topic of a text segment. We propose an unsupervised representation learning objective to capture the similar/different content selection preferences from the set of text segments by different news sources. 
    \item We conduct a case study on NELA-2020, a large-scale multi-national news corpus. We empirically validate the utility of our apporach, and discover that the derived representations of sources' selection preferences can be used as indicators for the ideological polarity among sources.  
\end{enumerate}



















\section{Background and Challenges}
\label{sec:background}
During news production, selection biases typically happen on the macro-level of editorial decisions, and thus it is widely believed that selection biases serve as good indicators for the implicit intent and motivating theme behind news sources. However, recognizing selection biases from news content alone is intrinsically difficult, as conceptually it requires knowing the \emph{population} of all the stories within a topic that news sources can cover, which \citet{d2000media} describe as ``unknowable and unidentifiable''.  
For the exact reason, past studies of media selection biases in communication research limit their scope on a focused event, such as political protests, where the complete set of newsworthy aspects can be reasonably estimated \cite{mccarthy1996images, hocke1996determining}. However, the general applicability of such findings is fairly limited, since sources do not typically conform to the same type of content selection decisions for different topics, as we also observe in this study.

From a pragmatic point of view, there exist two ways to discover the existence of selection bias from a source with respect to another. The first and most obvious way is through observations such as: \emph{Source A mentioned this fact or event, while Source B did not.} Other than the difficulty in assessing the truth value of such statements, such observations might often be attributed to less significant factors such as the mismatching size or publication frequency of the target news sources \cite{d2000media}. The second way, which we adopt in this study, is to identify cases where \emph{Source A and Source B discuss the same topics within a relatively small time frame, but they choose to mention them in very different context or aspects.} The right hand side of Figure~\ref{fig:lead_example} shows an example of such. Although such observation on its own does not reveal which side potentially exhibits selection bias, it represents the diverging interests of the sources on the matter. Along this line, our empirical results with such analysis across many sources reveal that the interests can be partially explained by the diverging ideological preferences of various sources. 

\paragraph{Challenge: Representing Topic} The first challenge of this space lies within the notion of \emph{topic}. The type of analysis we target in this study requires a fine-grained yet generally applicable definition of topic. The goal aligns well, at least conceptually, with the broad scope of computational frame analysis \cite{boydstun2013identifying}. \emph{Framing} generally refers to a set of nuanced discourse strategies to selectively report and highlight certain aspects or sub-topics over others when discussing a topic \cite{entman1993framing}. In practice, frame analysis usually starts from a pre-defined typology of sub-topic frames with respect to a set of controversial topics. Despite the recent efforts in extending the typologies and resources to increase coverage on topics and domains \cite{roy-goldwasser-2020-weakly}, the general applicability of such analysis remains limited. 

Our study aims to address the limitation of using a fixed frame typology by considering \emph{entities} as an alternative to sub-topic frame. This is in-part motivated by the fact that entities themselves usually provide strong topic guidance \cite{gabrilovich2007computing}. This enables us to align statements made by different sources on the same entity pairs, and study them comparatively. We will refer to such statements as \emph{context sentences} for a pair of entities throughout the rest of the paper. 

\paragraph{Challenge: Recognizing Selection Preferences}
The next piece of the puzzle is how we would represent the different content selection preferences of sources, given that set of statements they made containing the target pairs of entities. 

Formally, given two entities of interest $e_1$ and $e_2$, we want to learn a representation $g(C^s_{e_1, e_2} |  e_1, e_2)$ from a list of context sentences $C^s_{e_1, e_2}$ sampled from news articles by source $s \in S$. Again, context sentences $C^s_{e_1, e_2}$ are defined as all sentences that contain mentions of both $e_1$ and $e_2$ from source $s$. Collectively, $C^s_{e_1, e_2}$ represent all the statements that source $s$ made about $e_1$ and $e_2$ within the sample time frame. 

Given two sources $s_A$ and $s_B$, the conceptual goal is to identify whether $C^{s_A}_{e_1, e_2}$ are topically different from $C^{s_B}_{e_1, e_2}$. This objective in part resemble the family of conditional language modeling tasks \cite{keskar2019ctrl}, where the distributional divergence of the two sets of statement can be used as distance metric for their topical relatedness. In our case, we train a language model with the learning objective of recognizing whether two context sentences (with entity mentions masked) contain the same set of entities or not, with training data sampled from Wikipedia, We observe that language models learned through such objective is effective in capturing the high-level topical selection differences reflected in the context sentences.  

\section{Model Overview}
\label{sec:model}





\subsection{Extracting Entities As Topics}
\label{ssec:preprocess}
As a first step, we use entity linking and corpus statistics to identify and disambiguate a set of frequently mentioned entities within the NELA corpus as the topics of interest throughout our study. 
Due to the sheer scale of the corpus, we use the small version of NER model in spaCy \footnote{en\_core\_web\_sm} to identify named entity mentions. We opt for a simple yet effective method for entity linking \cite{ratinov2011local}. In Wikipedia, we compute the fraction of times a page of title $t$ is linked from another page with an anchor text/mention $m$. At inference time, given a mention $m$, we take the title with highest conditional probability, i.e. $\hat{t} = argmax_{t\in{T}}(P(t|m))$.

We then compute the frequency of the disambiguated entities in all NELA corpus and take the $1,000$ most frequent as the topics of interest for the study. 

\begin{figure*}[t]
\centering
\includegraphics[width=\linewidth]{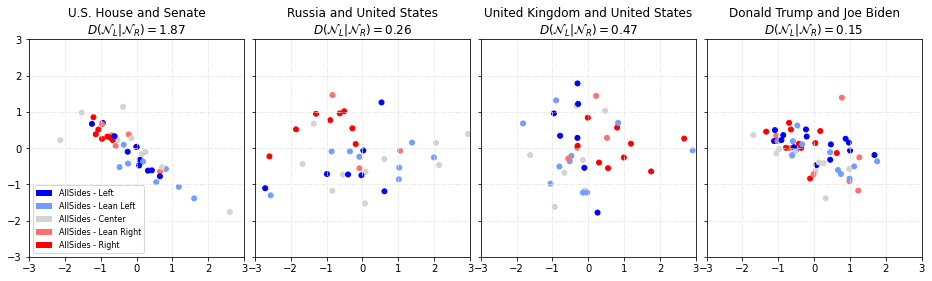}
\caption{PCA visualization of the source representations with respect to four representative pairs of entities. We consider the subset of 65 sources in NELA that are annotated by \emph{AllSides} with their political bias rating. Each dot in the plots represent a source, and is colored with respect to their politcal bias rating. $D(\mathcal{N}_{L}\|\mathcal{N}_{R})$ measures the \emph{left} vs \emph{right} distributional divergence of the source embeddings. More details can be found in \S~\ref{ssec:kl}}
\label{fig:scatter}
\end{figure*}

\subsection{Recognizing Topical Difference between Context Sentences}
\label{ssec:oprl}
We now describe the process of learning the topical representation of context sentence $c \in C^s_{e_1, e_2}$. Conceptually, as data sparsity can be a concern with entity pairs, we want the representation to capture the topical context of $e_1$ and $e_2$ described in $c$, while being agnostic to the topical information exposed by the surface form of $e_1$ and $e_2$ themselves.

Inspired by recent advances in neural knowledge base construction \cite{verga2021adaptable,Sun2021ReasoningOV}, we build an encoder model from pretrained \texttt{RoBERTa$_{base}$} model and parameters \cite{liu2019roberta}. For each input context sentence $c \in C^s_{e_1, e_2}$, we use a special token `\texttt{[ENT]}' to mask the mention of $e_1$ and $e_2$. In addition, we add `\texttt{[R1]}' and `\texttt{[R2]}' after the two entity masks respectively. The token embeddings of `\texttt{[R1]}' and `\texttt{[R2]}' would jointly represent the context in which the two entities get mentioned. The last hidden layer $h_{R1}$ and $h_{R2}$ of the two tokens are then forwarded through a linear layer to get the embedding $r_{e_1, e_2}$. 
\begin{equation*}
    r(c) = W^T [h_{R1}, h_{R2}]
\end{equation*}

To obtain training data for the language model, we sample all sentences with two or more mentions of the $1,000$ entities of interest from Wikipedia. This gives us about 3M sentences in total. We adopt triplet margin loss as our training objective. Each training instance consists of three context sentences -- For each anchor sentence $c \in C^s_{e_1, e_2}$, we randomly sample a positive sentence $c^+ \in C^s_{e_1, e_2}$ and a negative sentence $c^- \notin C^s_{e_1, e_2}$ which mentions at least one other entity other than $\{e_1, e_2\}$. The objective is to maximize the relative similarity between $r(c)$ and $r(c+)$, and we use L-2 norm as the distance function. 
\begin{gather*}
    \mathcal{L} = max[d(c, c^+) - d(c, c^-) + 1, 0] \\
    d(c_1, c_2) = \|r(c_1) - r(c_2)\|_{2}
\end{gather*}

Our model is trained on a single Nvidia TITAN RTX in 5 hours. We report the hyperparameter settings for the RoBERTa model in Appendix~\ref{appendix:hyperparam}.



Given the set of news articles published by each source $s \in \{s_1, ... , s_n\}$ in NELA, we identify all context sentences $C^s_{e_i, e_j}$ for each entity pair $(e_i, e_j)$. We run each context sentence $c \in  C^s_{e_i, e_j}$ through the context sentence encoder and mean-pool to get the result embedding. The embedding conceptually represents the collective content selection choices made by the source with respect to the target topic entities.  
\begin{equation*}
    g(C^s_{e_i, e_j}) = \frac{\sum_c r(c)}{|C^s_{e_i, e_j}|}
\end{equation*}

\section{Experiments and Results}
To assess the utility of our model framework, we investigate whether the learned representation of the sources' selection preferences correlate with the respective ideological preferences among sources. The main hypothesis we assess here is that sources with similar ideological preferences would discuss certain topics in more similar ways compared to sources with diverging interests. This way the distributional similarity/divergence of the source representations can be used as a tool to identify polarity among sources.  

For such, we focus on the 65 sources in NELA with 5-way political bias ratings from \emph{AllSides} (\{\emph{left, lean left, center, lean right, right}\}).

\subsection{Source Embedding Visualization}
\label{ssec:visual}
We first demonstrate that our langauge model representations of content selection preferences are indicative of the sources' selection biases and ideological preferences out-of-the-box. Figure~\ref{fig:scatter} shows the visualization of the first two principle components for the per-source embeddings of four example pairs of entities. We observe that the embeddings capture the following two factors of the typical context in which the two entities are described. 
\paragraph{Topic Variance}
Since the embeddings are learned through the conditional similarity between context sentences, naturally, the variance, or the ``scatteredness'' of the embeddings is indicative of the variety of topics in which the two entities are mentioned in. For example, as Figure~\ref{fig:scatter} shows, \textit{U.S. House and Senate} are generally more likely mentioned in similar context than \textit{Russia and United States}, regardless of the political tendency of the sources. 

\paragraph{Divergence among \emph{left} vs \emph{right} sources}
We observe that the representations of selection preferences for certain topics are directly indicative of the ideological divide between the  \emph{left} vs \emph{right} sources. From Figure~\ref{fig:scatter}, we see that \textit{U.S. House and Senate} is a good example of such.  For \textit{Russia and United States}, \textit{United Kingdom and United States}, despite the high topic variance, the PCA plot still suggests a clear divide between how the two sides typically frame the topic. We also observe counter-examples such as \emph{Donald Trump} vs. \emph{Joe Biden}, we suspect that as news mentioning the two in 2020 mostly focus on election, the two entities are more likely to be mentioned in similar context by different sources.

The findings suggest that we can recognize and characterize the polarity between sources, using a set of \emph{archetypical} entity pairs, as we will demonstrate in \S~\ref{ssec:prediction} and \S~\ref{ssec:understand}.

\begin{table}[]
    \centering
    \resizebox{1\linewidth}{!}{
    \begin{tabular}{ccccc}
    \toprule
    
Model  & Classifier & $P$ & $R$ & $F_{\beta=1}$ \\
\midrule
\multirow{4}{*}{RoBERTa} & Lasso Regression & 59.55 & 56.10 & 57.77 \\
 & Ridge Regression & 68.90 & 56.10 & 61.84 \\
 & Linear SVM & 76.51 & 56.10 & 64.73 \\
 & RBF-Kernel SVM & 70.49 & 68.29 & 69.37 \\
\midrule
\multirow{4}{*}{Ours} & Lasso Regression & 70.22 & 60.98 & 65.27  \\
 & Ridge Regression & 70.85 & 53.66  & 61.06 \\
 & Linear SVM & 74.49 & 60.98 & 67.06 \\
 & RBF-Kernel SVM & \textbf{80.49} & \textbf{75.61} & \textbf{77.97}\\
\bottomrule 
    \end{tabular} }
    \caption{
        Three-way polarity $\{$\textit{left, center, right}$\}$ prediction results on the MBFC test set, in weighted macro-averaged precision ($P$), recall ($R$) and $F_{\beta=1}$. We use the embeddings from 20 most frequent entity pairs in the NELA corpus as input features to the classifiers.
    }
    \label{tab:results}
\end{table}

\subsection{Source Polarity Prediction}
\label{ssec:prediction}
We now use a modeling approach to evaluate whether our language model representations of content selection preferences can be used as indicators for a sources' ideological preference.  We formulate our task as a three-way polarity prediction task: Given a set of $k$ entity pairs, we construct the representation $g^{s_i}$ of each source $s_i$ by concatenating $s_i$'s embeddings for all $k$ entity pairs. Then we train a classifier using $g^{s_i}$ as input features, and predict the three-way $\{$\textit{left, center, right}$\}$ political bias rating of the source $s_i$. 

\begin{table*}[t]
    \centering
    \resizebox{1\linewidth}{!}{
    \begin{tabular}{|cc|cc|}
    \toprule
    
Highest Divergence  & $D(\mathcal{N}_{L}\|\mathcal{N}_{R})$ & Lowest Divergence & $D(\mathcal{N}_{L}\|\mathcal{N}_{R})$\\

\cmidrule(lr){1-1} \cmidrule(lr){2-2}  \cmidrule(lr){3-3}  \cmidrule(lr){4-4} 

\textit{U.S. House of Representatives} and \textit{Senate} & 1.87 & \textit{Donald Trump} and \textit{U.S. House} & 0.07  \\
\textit{Iran} and \textit{United States} & 1.42 & \textit{United States} and \textit{U.S. Senate} & 0.07 \\
\textit{Donald Trump} and \textit{the F.B.I.} & 1.40 & \textit{Joe Biden} and \textit{Kamala Harris} & 0.07 \\
\textit{C.D.C} and \textit{United States} & 1.36 & \textit{United States} and \textit{Mexico} & 0.08 \\
\textit{China} and \textit{United States} & 1.12 & \textit{United States} and \textit{White House} & 0.08 \\
\textit{Donald Trump} and \textit{Russia} & 1.00 &
\textit{United States} and \textit{United States Congress} & 0.09 \\

\bottomrule 
    \end{tabular} 
    }
    \caption{
        Example entity pairs with the highest and lowest distributional divergence (\S~\ref{ssec:kl}) among sources with respect to their political bias ratings from \emph{AllSides}. 
    }
    \label{tab:kl}
\end{table*}

To construct a test set with sources not labeled by \textit{AllSides}, we collect the 5-way political bias ratings from Media Bias / Fact Check (MBFC)\footnote{\url{https://mediabiasfactcheck.com/}}. We aggregate the 5-way labels from both \textit{AllSides} and MBFC into 3-way $\{$\textit{left, center, right}$\}$. After removing the $54$ sources from MBFC that overlaps with AllSides, we use the rest $41$ sources as the heldout test set. We verify the label agreement between the two by evaluating the ground truth MBFC's labels against AllSides' using the overlap set, and get $81.99$ weighted macro-averaged $F_1$. More details can be found in Appendix~\ref{appendix:overlap}.

For baseline comparison, we replace our context sentence encoder with \texttt{RoBERTa}$_{base}$ without finetuning. As the main goal here is to  evaluate the descriptive power of the embeddings, we compare four types of popular linear/non-linear classifiers with relatively small parameter size to avoid overfitting, and freeze the embeddings from underlying embedding model during classifier training. For all models, we set $k=20$, and use the 20 most frequent entity pairs in the NELA corpus to construct the source representation and input features $g^{s_i}$ for each source $s_i$.

Table~\ref{tab:results} shows the evaluation results. We run 5-fold cross-validation on the \textit{AllSide} training set to select the hyper-parameters of the classifier models individually. We report the hyperparamter settings in Appendix~\ref{appendix:hyperparam}. We observe that the SVM model with RBF kernel, the only non-linear model, performs the best on the MBFC test set. More importantly, across all methods, except for Ridge Regression, our method achieved consistently better results than \texttt{RoBERTa}, suggesting the benefit and advantage of our learning objectives on modeling sources' selection biases.

\section{Analysis}

\subsection{Identifying Archetypical Entity Pairs for Ideological Divergence}
\label{ssec:kl}
In the previous section, we empirically show that, depending on the choice of entity pairs, the content selection preferences can be indicative of the divide between \emph{left vs. right} ideological preferences. Conceptually, identifying such a \emph{archetypical} set of entity pairs would provide us an interface to better recognize and characterize polarity among sources, and in turn enable us to reason about their interests and patterns with respect to the each topic.

We now describe a statistical procedure to automatically identify such \emph{Archetypical Entity Pairs (AEPs)} by measuring the distributional divergence of the embeddings from \textit{left} vs \textit{right} sources. 
Like \S~\ref{ssec:prediction}, we aggregate five-way media bias ratings from \textit{AllSides} into three-way $\{$\textit{left, center, right}$\}$ labels. For each entity pair $\{e_i, e_j\}$, we use the distributional divergence, measured by two-way Kullback–Leibler (KL) divergence, of left- vs. right-leaning sources to measure whether $\{e_i, e_j\}$ is a good indicator for the divide between \textit{left} vs. \textit{right} sources. For such, we fit multi-variate Gaussian distribution $\mathcal{N}_{L}(\mu_{L}, \Sigma_{L})$ and $\mathcal{N}_{R}(\mu_{R}, \Sigma_{R})$ to the $d$ dimensional embeddings of left- vs. right-leaning sources respectively. We then compute the KL-divergence between the two Gaussians.
\begin{gather*}
\label{eq:kl}
    D(\mathcal{N}_{L}\|\mathcal{N}_{R}) = \frac{1}{2} (tr(\Sigma_R^{-1}\Sigma_L) + ln\frac{|\Sigma_R|}{|\Sigma_L|} \\ + (\mu_R - \mu_L)^T\Sigma_R^{-1}(\mu_R - \mu_L) - d) 
\end{gather*}

We compute $D(\mathcal{N}_{L}\|\mathcal{N}_{R})$ for the 100 most frequent entity pairs in the NELA corpus. The average $D(\mathcal{N}_{L}\|\mathcal{N}_{R})$ is $0.386$, with standard deviation of $0.394$. Table~\ref{tab:kl} shows the entity pairs with the highest and lowest divergence respectively. 

Conceptually, this provides us a systematic way to quantify which entity pairs serve as better indicators for the divide between the content selection preferences from groups of sources with different political tendencies.

\subsection{Understanding Sources' Selection Biases with Archetypical Entity Pairs}
\label{ssec:understand}
In \S~\ref{ssec:kl}, we demonstrated a way to construct a set of AEPs through distributional divergence measurements.   
We now demonstrate, through a case study, how a such set of AEPs can be used to characterize sources' selection biases. 

We start by selecting five example sources of interest: \textit{Breitbart} (Right), \textit{Fox News} (Lean Right), \textit{CNN} (Center), \textit{The New York Times} (Lean Left), and \textit{The Huffington Post} (Left). 
We select the 10 entity pairs with the highest $D(\mathcal{N}_{L}\|\mathcal{N}_{R})$, that also have more than $10$ context sentences present in the corpus. 
For each entity pair, we train an individual linear SVM classifier\footnote{We choose not to use the best performing RBF-kernel SVM here, as it is more sensitive to hyperparamter choices, which are more difficult to tune for in this setting.} 
to predict the three way political bias rating of the sources using the AllSides training set, with the five sources of interest held out from training. We use the individual SVM classifiers to classify the political bias rating of the five sources with respect to each entity pair. 

\begin{figure}[h]
\centering
\includegraphics[width=\linewidth]{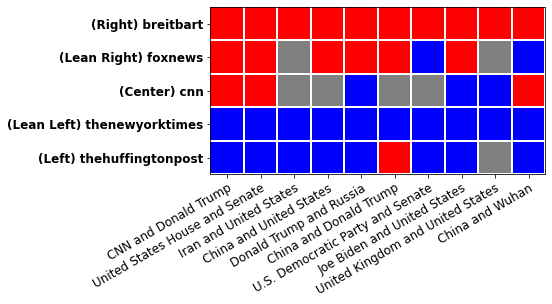}
\caption{Predicted polarity \{\textit{\textcolor{blue}{left}, \textcolor{gray}{center}, \textcolor{red}{right}}\} of five held-out sources (with most prolific publication rates in the NELA corpus) of different political bias ratings with respect to ten archetypical entity pairs (AEPs) with highest $D(\mathcal{N}_{L}\|\mathcal{N}_{R})$ values. We observe that the distribution of predicted polarity on the AEPs is indicative of the sources' political bias ratings. Some entity pairs shown in Table~\ref{tab:kl} are omitted from this figure, due to 
lack of coverage by one or more of the five sources. 
}
\label{fig:heatmap}
\end{figure}

Figure~\ref{fig:heatmap} shows the prediction results. We observe that label distribution on the 10 entity pairs indicates the political bias rating of the individual sources. In the case of \textit{The New York Times} and \textit{Breitbart}, the predicted polarity on all 10 entity pairs consistently reflect their political bias ratings. Some disagreements can be observed with the other three sources. Interestingly, in the case of CNN, the predicted polarity on the 10 entity pairs are mixed, potentially indicating the source's more centered or unbiased position on the political bias spectrum.

With such analysis, we reach the interesting observation that sources don't conform to their \emph{perceived} ideological preference or political tendency on every matter. Depending on the topic, the same source could very well exhibit opposite attitude to their usual ideological preference. Similarly, sources with different ideological preferences do not always disagree with each other, and there exist common grounds. We argue that the key advantage of our proposed methodology is that it can uncover such agreements/disagreements on a fine-grained topic level. This allows us to jointly reason and characterize the selection biases and (political) polarity in an interpretable way.

\subsection{Probing with Subjective Predicates Templates}
\label{ssec:sentiment}
To understand what media attitude or sentiment that the context sentence embeddings imply or entail, we design a probing method to our embeddings that leverages the semantics of subjective predicates. 

\begin{figure}[t]
\centering
\includegraphics[width=\linewidth]{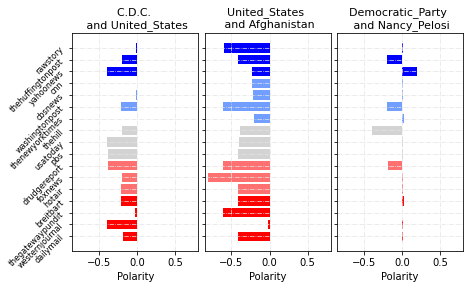}
\caption{Sentiment polarity score (\emph{positive}$\rightarrow$1 and \textit{negative}$\rightarrow$-1) by a list of sources for three example pairs of entities, inferred from the projected similarity between subjective verb predicates (\S~\ref{ssec:sentiment}) and the context sentence embeddings by each source. The sources are colored according to their political bias rating from \emph{AllSides}. }
\label{fig:polarity}
\end{figure}

Following similar strategies proposed by \citet{iyyer2016feuding} and \citet{han2019no}, we 
construct a list of subjective predicates and project our embeddings as a distribution over the subjective predicates. For such, we sample 400 verb predicates with its sentiment polarity labels (\textit{positive} vs \textit{negative}) from the MPQA subjective lexicon \cite{wilson2005recognizing}, with the 200 most frequent positive and negative polarity verb predicates in the NELA corpus respectively. With each verb predicate, we formulate a prompt in the same formats as the inputs to our context sentence encoder (\S~\ref{ssec:oprl}): We prepend \texttt{[ENT][R0]} and append \texttt{[ENT][R1].} to the verb predicate as a probing prompt to the model. We encode the probing prompt for each individual verb predicate using the context sentence encoder. Given an context sentence embedding from a source, we compute its cosine similarity to the embeddings of all 400 verb predicate prompts. We take the top 10 most similar verb predicates, and take the weighted mean of the polarity label (\textit{positive} $=1$ vs \textit{negative} $=-1$) of the verb predicates as the sentiment score for the context sentence. The weight on each verb predicate is determined by the softmax probabilities over the cosine similarity scores.   

Figure~\ref{fig:polarity} shows the projected sentiment score on three example entity pairs from a list of sources with different political bias ratings. We observe that in the cases of \textit{C.D.C} and \textit{United States}, as well as \textit{Afghanistan} and \textit{United States}, sources from different sides mostly exhibit negative sentiment when describing the entity pairs. In the case of \textit{Afghanistan} and \textit{United States}, the overall negative sentiment is stronger, and more conservative sources seem to exhibit more polarized attitude compared to the liberal sources. Similar observations can be made on \textit{C.D.C} and \textit{United States}. In the case of \textit{Democratic Party} and \textit{Nancy Pelosi}, we see a more mixed sentiment from most sources, with only a few exceptions. 

The probing result shows once again the nuanced difference in how sources typically frame different set of topics, and how such difference does not necessarily conform with their perceived political bias.
\section{Related Work and Discussion}
\label{sec:related_work}
In this study, we propose and study a conceptual entity-focused approach to represent and reason about the similarity/difference of various sources' content selection preferences. We in turn discover that such representations can be used as describe their ideological polarity. 

Besides the use case in this study, named entities have many other implications to discourse modeling and understanding in NLP.  \citet{choi2016document} propose a general ILP-based inference method for tracking the dynamic sentiment relations between entities mentioned in the same document. \cite{iyyer2016feuding} introduce a unsupervised learning method to project the relation between fiction characters on to word embedding space. \citet{han2019no} adopt a similar strategy to track the dynamic relationship between nations, as depicted by news media. \citet{you2020relationship} formulate a cloze style task for predicting the relation between two fictional characters in the multiple-choice question-answering format. 

The goal of our study aligns in part with frame analysis. The goal of frame analysis is typically identifying selection biases, manifested by frame usages in articles, and study its connection to certain attributes of information sources, such as ideological preferences. Frame analysis usually starts from a pre-defined typology of sub-topic frames with respect to a set of controversial topics. \citet{boydstun2013identifying} defines a set of 15 commonly used frames by news media,  and \citet{card2015media} introduces a large-scale corpus, where documents of three topics are hand labeled with their sub-topic frame usage. A series of studies has since used the two resources to develop classifiers for predicting the primary frame of a document \cite{card2016analyzing, ji2017neural, field2018framing}. Our work instead looks to alleviate the limits of a predefined typology, and instead use entities as an open-domain way to represent topic. 

It's worth noting that the goal of this work is intrinsically different from the political polarity prediction line of work \cite{preoctiuc2017beyond, kulkarni2018multi}, as our goal is not trying to predict political polarity for the sources, but instead using polarity as an interface to characterize the sources' selection biases. 


The motivation of our work is closest in spirit to a recent study by \citet{pujari-goldwasser-2021-understanding}, where they represent the political figures with their statements with regards to different entities. However, their approach requires a predefined set of political figures, news events, entities, which limits scalable analysis on more timely news content. As we design our context sentence encoder to be entity-agnostic, our method does not have such requirements. Our framework instead leverages the coverage on entity pairs that can be aligned across multiple sources. Although such formulation requires no labeled data, data scarcity can be a concern due to the sparse nature of entity pairs.

\section{Conclusion}
In this paper, we discuss and investigate the challenges of discovering patterns of selection biases from a massive-scale news corpus. Through a case study, we empirically validate the hypothesis that a source's selection biases have correlation with its ideological leaning.  


With our empirical experimental results, we reach the interesting observation that sources do not necessarily conform to the same expected way of framing topics, and sources with different ideological preferences would still have common ground when it comes to certain topics. Therefore, we argue that understanding the selection biases in news media requires far more than assigning labels on the source-level. Instead, a good, alternative strategy is to comparatively analyze the statements made with respect to various topics among different sources, and understanding the implications of such difference in connection to the ideological/political preferences of the sources. Our paper merely offers one speculative way of doing so, and We hope that our proposed methodology and findings will facilitate future work on alone this line. 

\section*{Limitations}
Since the nature of the paper is provide initial speculative insights and empirical validation of our conceptual model, our method, in its current form, has many limitations. For one, even though our method requires no labeled data, it indeed requires a large amount of unlabeled data (for enough entity coverage). Specifically, our analysis method requires enough presence of context sentences per entity pair in the corpus. 

As mentioned in $\S~\ref{sec:background}$, our study only focus on comparing the topics that both/all sources cover. We leave the more challenging cases of partial source coverage for future work. 

It's worth noting that the analysis tool developed in this study only provides speculations for studying selection biases and polarity of news sources. The results alone do not stand as evidence for such. 

\section*{Acknowledgements}
This work was supported by a Focused Award from Google, and by Contract FA8750-19-2-1004 with the US Defense Advanced Research Projects Agency (DARPA). Approved for Public Release, Distribution Unlimited. The views expressed are those of the authors and do not reflect the official policy or position of the Department of Defense or the U.S. Government.

\bibliography{anthology,new, ccg, cited_10s}
\bibliographystyle{acl_natbib}

\newpage
\appendix

\section{Models and Hyperparamter Settings}
\label{appendix:hyperparam}
We implement our context sentence encoder (\S~\ref{ssec:oprl}) using Huggingface's \texttt{transformers} library \cite{wolf-etal-2020-transformers}, under Apache 2.0 license. We use and initialize the parameters of our models with pretrained \texttt{roberta-base} model.   

We train the context sentence encoder with ADAM optimizer, with learning rate $=1e-5$ and $3$ training epoch. Our model is trained on a single Nvidia TITAN RTX in 5 hours.

For the classifiers, we use \texttt{scikit-learn}'s implementation of the four classifier models. Below we describe the hyperparamter settings of each model individually.  
\begin{enumerate}
    \item Lasso: $C=1$
    \item Ridge: $\alpha=1.0$ 
    \item Linear SVM:  $C=0.1$ with L-2 penalty and hinge loss
    \item RBF-Kernel SVM: $C=2$, with polynomial degree=3 for the kernel function. $\gamma=0.01$ 
\end{enumerate}
The weights for all classifiers are zero initialized. 
\section{Label Agreement between \textit{AllSides} and MBFC}
\label{appendix:overlap}
MBFC has $54$ sources labeled with 5-way political bias ratings that overlaps with AllSides annotations. We verify the label agreement between the two by evaluating the ground truth MBFC's labels against AllSides' using the overlap set, and get precision=$84.50$, recall=$79.63$, and $81.99$ weighted macro-averaged $F_1$. We show the confusion matrix between the overlap set below.

\begin{figure}[h]
\centering
\includegraphics[width=0.6\linewidth]{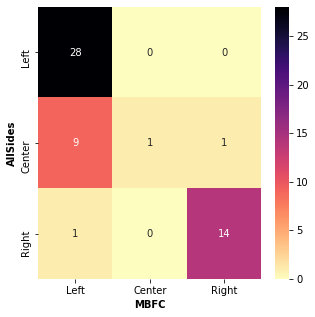}
\caption{Confusion matrix of label agreement between MBFC and AllSides on the overlapping set of 54 sources. }
\label{fig:polarity}
\end{figure}

\end{document}